\pdfoutput=1

\documentclass[11pt]{article}

\usepackage[]{EMNLP2023}

\usepackage{times}
\usepackage{latexsym}

\usepackage[T1]{fontenc}

\usepackage[utf8]{inputenc}

\usepackage{microtype}

\usepackage{inconsolata}

\usepackage{microtype}
\usepackage{hyperref}
\usepackage{url}
\usepackage{booktabs}
\usepackage[]{mdframed}
\usepackage{framed}
\usepackage{amssymb}
\usepackage{pifont}
\usepackage{xspace}
\usepackage{graphicx}

\usepackage{multirow}

\newcommand{\ultra}{\texttt{Gemini-Ultra}\xspace}
\newcommand{\pro}{\texttt{Gemini-Pro}\xspace}

\newcommand\blfootnote[1]{%
  \begingroup
  \renewcommand\thefootnote{}\footnote{#1}%
  \addtocounter{footnote}{-1}%
  \endgroup
}

\title{Enhancing Incremental Summarization with Structured Representations}

\author{Eunjeong Hwang$^{\dagger}$\\
University of British Columbia\\
\texttt{ejhwang@cs.ubc.ca} \\\AND
Yichao Zhou, James Bradley Wendt, Beliz Gunel, Nguyen Vo, Jing Xie, Sandeep Tata \\
Google Deepmind \\
\texttt{\{yichaojoey, jwendt,  bgunel, nguyenvo, lucyxie, tata\}@google.com}
}

\begin{document}
\maketitle
\blfootnote{$^{\dagger}$This work was completed while the author was working as an intern at Google Deepmind.}
\begin{abstract}

Large language models (LLMs) often struggle with processing extensive input contexts, which can lead to redundant, inaccurate, or incoherent summaries. Recent methods have used unstructured memory to incrementally process these contexts, but they still suffer from information overload due to the volume of unstructured data handled. In our study, we introduce structured knowledge representations (GU\textsubscript{json}), which significantly improve summarization performance by 40\% and 14\% across two public datasets. Most notably, we propose the Chain-of-Key strategy (CoK\textsubscript{json}) that dynamically updates or augments these representations with new information, rather than recreating the structured memory for each new source. This method further enhances performance by 7\% and 4\% on the datasets.

\end{abstract}

\section{Introduction}
\label{sec:intro}

Individuals commonly use large language models (LLMs) to summarize content from sources like webpages, books, and articles \cite{jin2024comprehensive,kryscinski2021booksum,SciSumm,gunel2024strumllm}. This aids in efficiently processing large volumes of information, influencing daily decision-making tasks. Despite their potential, LLMs often struggle with processing extensive contexts, leading to redundancy or inaccuracies~\cite{hwang2024sumie}. Recent research integrates unstructured memory systems~\cite{madaan2022memory, zhang2023memory} and fine-tunes models for larger context windows~\cite{long-context}. However, unstructured memory formats often result in oversized memories that overload the model, impairing its processing and summarization abilities. \citet{zhang2023memory} proposes a self-controlled memory architecture to manage information via heuristics, yet unstructured memory complicates retrieval, and \citet{li2024longcontext} finds that larger context windows still struggle to process documents comprehensively.

\begin{figure}[t]
    \centering
    \includegraphics[width=\linewidth]{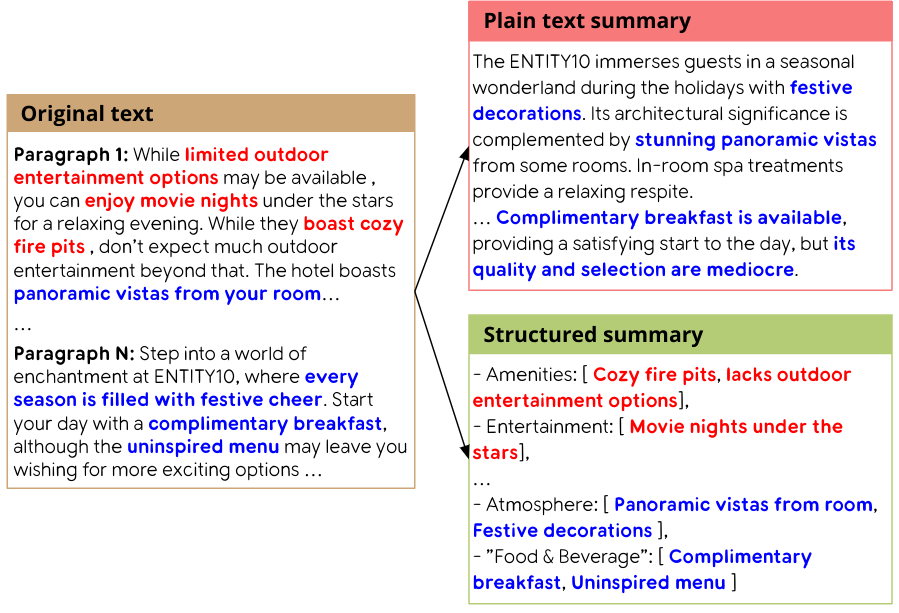}
    \caption{Example summaries generated based on a text memory representation and a structured memory representation, respectively, given the N paragraphs in the original texts. Red color marks the values that are not included in plain text summary.}
\label{fig:motivation}
\vspace{-10pt}
\end{figure}



To address summarization tasks over extensive input contexts, we introduce CoK, the Chain-of-Key update based on JSON structured memory representations. On one hand, this approach leverages two key aspects: (1) JSON's capability to organize information into distinct, easily accessible segments, facilitating efficient updates and retrievals--offering superior expansion and nesting capabilities compared to markdown tables or graphs~\cite{Dagdelen2024}; (2) The prevalence of JSON in LLM pretraining data enhances LLMs' ability to understand and generate structured JSON content~\cite{openai,xia2024fofo}\footnote{Our method uses the JSON format, though YAML or XML could also serve as structured formats.}. Consequently, as shown in Figure~\ref{fig:motivation}, JSON-based representations produce more well-structured summaries than plain text, avoiding verbosity and retaining critical content across all sections.

On the other hand, instead of requiring LLMs to recreate complete structured representations for all previously encountered knowledge upon encountering a new information source, our method dynamically identifies which new information needs to be added or updated within the existing structures. This reduces the cognitive load on LLMs, enabling them to maintain and process complex information more effectively and produce more accurate and comprehensive summaries, as illustrated in Figure~\ref{fig:motivation}, where summaries using the CoK strategy effectively retain essential knowledge, unlike unstructured summaries that often miss key details (highlighted in red).


\paragraph{Contributions:}
\begin{itemize}
    \item \vspace{-3pt} We demonstrate that structured knowledge representations significantly enhances LLMs' summarization capabilities, improving performance by 40\% and 14\% on two public summarization benchmarks.
    \item \vspace{-3pt} We introduce CoK, the Chain-of-Key update strategy, which dynamically updates or augments structured representations with new information, boosting performance further by 7\% and 4\% on the benchmarks, without needing to recreate the JSON structure for each new source.
    \item \vspace{-3pt} We offer an analysis demonstrating that structured representations enable models to retain more relevant contexts and historical information than plain-text memory, particularly when token availability for storing information is limited.
\end{itemize}


\begin{figure}[t]
    \centering
    \scriptsize
    \includegraphics[width=.9\linewidth]{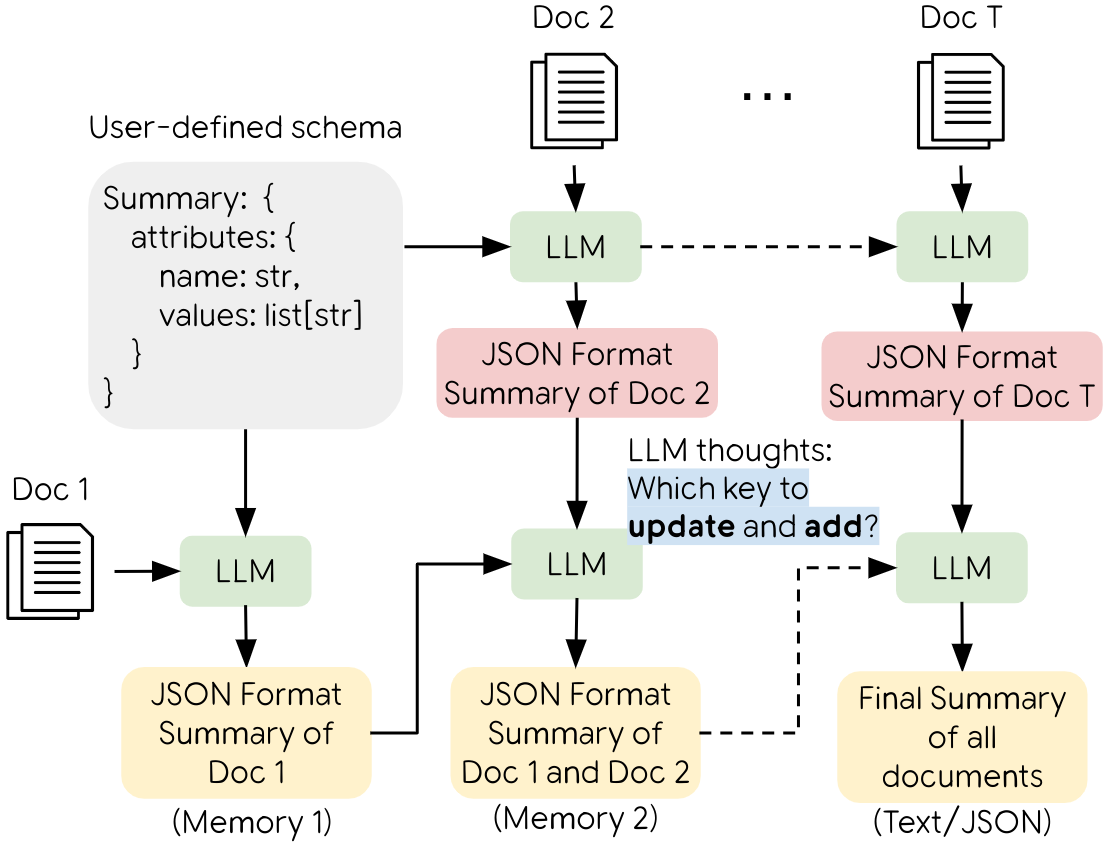}
    \caption{Overall incremental summarization process with a structured memory representation based on user-defined schema. The questions under ``LLM thoughts'' are for Chain-of-Key Updating process.}
    \label{fig:overall-architecture}
    \vspace{-10pt}
\end{figure}

\section{Methodology}
\label{sec:method}

\paragraph{Overview.}

The goal of the summarization task is to generate a summary $S_t$ from a sequence of documents $D = {D_0, D_1, \dots, D_t}$ that encapsulates the essential contents of all documents up to time $t$. We primarily approach the task within an incremental framework: \textbf{Incremental Summarization}. This process involves continuously refining the summary by integrating new information from each subsequent document. Each document $D_i$ may contain new, relevant information that contributes to an ongoing topic. The task requires producing a summary $S_t$ at each time step $t$, encompassing all critical information accumulated, thereby reflecting the key insights from the document sequence. Our methodology utilizes a structured representation to efficiently manage and update the summary dynamically with each new document.

\paragraph{Initial Structured Summary Generation.}
As depicted in Figure~\ref{fig:overall-architecture}, we initiate the process by establishing a schema tailored to the task. This schema is then provided to the LLM along with any newly available information and, if applicable, any pre-existing data in memory. The LLM is tasked with generating outputs that align with the key elements specified in the schema.



\paragraph{Chain-of-Key: Structured Summary Updates.}
We propose the Chain-of-Key (CoK) updating method to merge existing memory data with new input in a structured summary. First, the LLM creates a structured summary from a new document. This new summary, together with any existing structured memories, is then processed by the model. The method uses two main operations, \texttt{Update} and \texttt{Add}, to effectively combine the summaries.

The \texttt{\textbf{Update}} operation modifies existing summary to incorporate new data. It progresses through three steps: (1) identifying keys in the existing summary, such as [\texttt{``Amenities'', ``Food \& Beverage''}], (2) matching these keys with relevant ones in the existing memory, like \texttt{``Amenities''}, and (3) generating a JSON path to integrate the new key information, such as \texttt{\$.attributes.Amenities}. This path is then used to programmatically update the summary.

The \texttt{\textbf{Add}} operation focuses on incorporating new keys from the new information that are absent in the existing summary.
It also follows three steps: (1) identifying new keys in the summary, such as [\texttt{``Noise Level''}], (2) detecting which of these are not in the existing summary, like \texttt{``Noise Level''}, and (3) generating a JSON path for adding the new key, such as \texttt{\$.attributes.Noise Level}. Using the generated path, new values are programmatically added to the summary.

Breaking down the updating process into sub tasks employs the LLM's reasoning capabilities to tackle complex and multifaceted reasoning problems in the incremental summarization task. See Appendix \ref{app:schema} and \ref{app:cok-prompts} for the schemas and prompts.

\paragraph{Final Summary Generation.}
Once the iterative summary update process at each step is completed, the LLM receives the aggregated memory from $D_0, D_1, ..., D_{t}$ and generates the final summary $S_t$. This summary can be presented in various formats, such as JSON or plain text.

\section{Experimental Setup}
\label{sec:exp_setup}

\paragraph{Dataset.} We evaluate our methods using two datasets: SUMIE~\cite{hwang2024sumie} and BooookScore~\cite{chang2024booookscore}. The SUMIE dataset is designed to assess the incremental entity summarization capabilities of LLMs. The BooookScore dataset is aimed at long document summarization and includes 100 recently published books, with some books exceeding 100k tokens. Due to our models' 6K token context window size, each book is segmented into 2K-token chunks. For additional dataset details, see Appendix~\ref{app:dataset}.


\paragraph{Baseline.} 
We compare our method against three setups: \texttt{Generate-Once} (GO), \texttt{Generate-Update} (GU), and \texttt{Generate-Merge} (GM), using two data formats—JSON and plain text—with two state-of-the-art LLMs: Gemini-Ultra and Gemini-Pro\footnote{\url{https://deepmind.google/technologies/gemini/}\\Model temperatures are all set to 0.8 by default.}. In GO, the LLM generates a comprehensive summary from all related paragraphs in a single step. In GU, the LLM incrementally generates updated summaries by integrating each new paragraph. In GM, the LLM merges summaries from each new paragraph incrementally, utilizing JSON for its key-matching capability to facilitate merging. Programmatic merging in JSON may retain redundant key-values, which are removed by directing the LLM to filter out such redundancies. Details on the prompts for these methods are in Appendix \ref{app:sumie-basline-prompts} and \ref{app:book-basline-prompts}.

\paragraph{Evaluation Metrics.} 
For SUMIE, we employ its LLM-assisted evaluation method to measure precision, recall, and F1 scores of the final summary. For the BookScore dataset, we utilize its LLM-based metric to assess summary coherence, evaluating across eight predefined error dimensions: entity omission, event omission, causal omission, discontinuity, salience, language inconsistency, and duplication. See Appendix~\ref{app:eval-details} for more setup details.

%

\paragraph{Limited Token Scenario.} To evaluate how much information JSON and text formats retain in in-context memory, we established a scenario with a constrained in-context memory token limit of $K$ tokens. This constraint is crucial for handling long documents, like books, that exceed the model's context window. For the SUMIE dataset, $K$ is set to 200 and 300 tokens, while for the BooookScore dataset, it's set to 1000 tokens. See Appendix~\ref{app:compress-prompts} for more details about compression criteria and the associated prompts.


\section{Results}
\label{sec:results}

\begin{table}[t]
\scriptsize
\centering
\begin{tabular}{lr|rrr|rrr}
\toprule
 &  &  \multicolumn{3}{c}{Ultra} & \multicolumn{3}{c}{Pro} \\ 
 & Turn & P & R & F1 & P & R & F1 \\ 
\midrule
GO\textsubscript{text} & last & 86.2 & 42.7 & 56.4 & 85.8 & 40.4 & 54.1 \\ 
GO\textsubscript{json} & last & 91.2 & 58.9 & 70.9 & 85.1 & 61.0 & 70.1  \\ 
\midrule
GU\textsubscript{text} & start & 77.3 & 70.0 & 72.6 & 74.7 & 66.9 & 69.7  \\ 
 & last & 76.6 & 45.2 & 55.8 & 73.3 & 26.8 & 38.4 \\ 
 & Avg. & 76.1 & 54.3 & 62.2 & 73.0 & 41.8 & 51.2 \\ 
\midrule
GU\textsubscript{json} & start & 88.6 & 81.6 & 84.3 & 85.1 & 80.7 & 82.2  \\ 
 & last & 80.2 & 76.7 & 78.1 & 81.7 & 69.4 & 74.7 \\ 
 & Avg. & 80.9 & 78.9 & 79.4 & 83.4 & 74.0 & 77.9 \\ 
\midrule
GM\textsubscript{json} & start & 88.6 & 82.9 & 85.0 & 84.2 & 82.6 & 82.6 \\ 
 & last & 86.8 & 63.2 & 72.7 & 84.6 & 74.3 & 78.6 \\ 
 & Avg. & 86.5 & 70.9 & 77.3 & \textbf{84.7} & 78.7 & 80.9 \\ 
\midrule
CoK\textsubscript{json} & start & 89.1 & 77.8 & 82.8 & 81.1 & 80.5 & 79.9 \\ 
 & last & 92.6 & 78.0 & 84.5 & 84.6 & 83.6 & 83.9 \\ 
 & Avg. & \textbf{91.8} & \textbf{80.5} & \textbf{85.5} & 83.7 & \textbf{83.9} & \textbf{83.5} \\ 
\bottomrule
\end{tabular}
\caption{Overall performance of Ultra and Pro models on the SUMIE dataset. "start" indicates performance at the first paragraph, and "last" represents performance at the last paragraph aggregating all attribute-value pairs. P, R, and F1 refer to the average precision, average recall, and macro F1 scores, respectively.}
\label{tab:eval-results}
\vspace{-20pt}
\end{table}

\paragraph{Text vs. JSON, Table \ref{tab:eval-results}, \ref{tab:eval-results-book}.} 
Table \ref{tab:eval-results} shows that the JSON format outperforms plain text in the incremental summarization tasks using the Ultra and Pro models, with notable differences in both the GO and GU methods. Specifically, GO\textsubscript{json} averages a 28\% F1 score improvement over its text equivalent, while GU\textsubscript{json} sees a 40\% improvement. This discrepancy primarily stems from the low recall with the text format, suggesting that plain text leads to information loss over iterations. The JSON format, however, supports better information retention. This is evident in later iterations and the book summarization task shown in Table~\ref{tab:eval-results-book}. Here, GU\textsubscript{json} posts a 14\% gain, enhancing the model's ability to maintain key details about characters and events.

\begin{table}[t!]
\scriptsize
\centering
\begin{tabular}{l|cccc}
\toprule
 Model & GU\textsubscript{text} & GU\textsubscript{json} & GM\textsubscript{json} & CoK\textsubscript{json} \\
\midrule
Pro & 53.1 &  58.5 & 61.5 & \textbf{62.2} \\
 Ultra & 51.9 &  61.7 & 60.1 & \textbf{63.1} \\

\bottomrule
\end{tabular}
\caption{BookScore performance on GU, GM, and CoK, where the token size for existing information was limited to 1000 tokens.
}
\label{tab:eval-results-book}
\vspace{-10pt}
\end{table}

\paragraph{Effectiveness of Chain-of-Key Update, Table \ref{tab:eval-results}, \ref{tab:eval-results-book}.} 


Table \ref{tab:eval-results} illustrates the effectiveness of the Chain-of-Key (CoK) method, which significantly outperforms all baseline models. Specifically, the CoK method, when applied with the Pro model, surpasses the best JSON baseline (GM\textsubscript{json}) of the larger Ultra model on the SUMIE task by 10\% in F1 score. Additionally, CoK achieves a 7\% F1 improvement over the GU\textsubscript{json} and GM\textsubscript{json} methods, averaged over Pro and Utrla models.

The analysis also highlights a notable decrease in recall for the GM method after removing duplicates over turns. This is more pronounced in the Ultra model, which removes more attribute-value pairs than the Pro model, leading to lower recall. In contrast, the CoK approach enhances both precision and recall across turns in both models, improving the F1 score by 3\% in the final turn. This improvement suggests that CoK's step-by-step processing allows the model to more accurately select and update information, maintaining relevance as iterations progress.

Further validation comes from Table~\ref{tab:eval-results-book}, where CoK shows 3\% and 4\% improvements in book scores over GM and GU. This indicates CoK's effectiveness in preserving detailed explanations of complex entities and events within books, crucial for the narrative. Although current metrics do not measure recall in book scores, this highlights an area for future research.

\begin{figure}[t]
    \centering
    \small
    \includegraphics[width=0.75\linewidth]{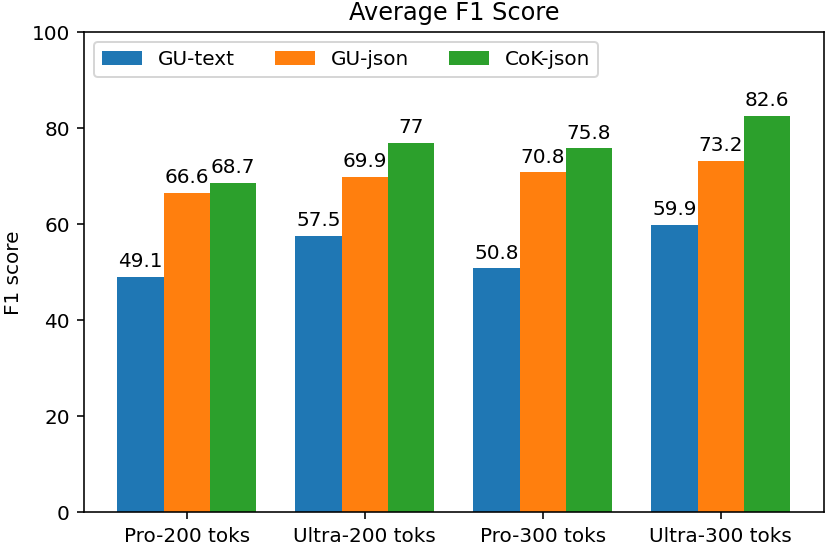}
    \caption{Average F1 score across all turns with limited memory token size on SUMIE.}
    \vspace{-10pt}
\label{fig:sumie-compress}
\end{figure}

\paragraph{Limited in-context token size for existing information, Figure \ref{fig:sumie-compress}, Appendix \ref{app:sumie-compress-tokens}.}
The CoK method uses significantly more tokens (an average of 604 tokens) for in-context memory compared to text baselines (an average of 269 tokens), raising the question of whether JSON can hold more information than text within the same token constraints. As shown in Figure~\ref{fig:sumie-compress} and detailed in Appendix~\ref{app:sumie-compress-tokens}, both the average F1 scores and the number of tokens used as existing information are tracked across all turns when token size is limited. The GU\textsubscript{json} method substantially surpasses baseline methods even with restricted token counts, achieving a 30\% average F1 score improvement over textual counterparts. CoK shows an extra 8\% F1 score improvement, suggesting that JSON format maintains more precise and distinct information in summaries.


\paragraph{Error Case Study.}
We observe that structured memory methods often add excessive details to book summaries. Here are sentences from two different summaries generated by CoK:

\begin{mdframed}[font=\scriptsize]
1. ... The family received support from extended family members and healthcare professionals, including \textcolor{red}{Katie}, \textcolor{red}{Angela}, \textcolor{red}{Rachel}, and \textcolor{red}{Mira}. \textcolor{red}{Lola}, a therapy dog, brought joy during Henry's illness. ... \\
2. ...Eleanor Bennett's children, \textcolor{red}{Benny}, \textcolor{red}{Byron}, and \textcolor{red}{Marble}, are grappling with their complex family history and personal struggles....
\end{mdframed}
The red text highlights unnecessary details, like minor character names, retained for broader coverage. This discrepancy underscores a gap in the LLM’s approach to what constitutes a comprehensive summary versus an effective book summary. While structured representations help retain more details, this excess negatively affects two evaluation metrics of book score: entity omission (mentioning entities without desciptions) and salience (including trivial details irrelevant to the storyline). Managing the level of detail in structured summaries poses a significant challenge for future research.

\section{Conclusion and Discussion}
\label{sec:conclusion}

In this paper, we introduce the Chain-of-Key method, which uses structured memory representations and leverages LLM's step-by-step reasoning to dramatically improve performance on two summarization tasks, surpassing strong baselines. 
JSON demonstrates its superiority in organizing knowledge for incremental summarization\footnote{See Appendix~\ref{app:markdown} for a comparison between JSON and Markdown tables, discussing the unique capabilities of JSON.}.
Challenges remain in filtering out trivial details and focusing on crucial information within structured summaries. Developing heuristics based on structured properties to better highlight key information is an area for future research.

\section*{Limitations}
\label{sec:limitations}

Our approach capitalizes on the inherent capabilities of LLMs to generate structured JSON formats. However, while most recent LLMs manage this well, smaller models such as Llama3-8B, Mistral-7B, and Gemini Nano often produce structured outputs with errors. 

\noindent In terms of evaluation, we adhere to the methods outlined in SUMIE and BookScore, which rely on LLM-based metrics. These evaluations are both computationally intensive and time-consuming. 

\noindent Additionally, although our method improves recall in final summaries, the book summarization task currently lacks a specific metric for measuring recall. We also did not evaluate the redundancy and accuracy of the information produced by the LLMs.

\section*{Ethics Statement}
\label{sec:ethics}
The LLMs we used to evaluate are trained on a large-scale web corpus and may also bring some bias when generating sentences (or structured data) or when evaluating final summries of entities or books. We evaluated our method on publicly available datasets.

\bibliography{anthology,custom}

\begin{thebibliography}{13}
\expandafter\ifx\csname natexlab\endcsname\relax\def\natexlab#1{#1}\fi

\bibitem[{Agarwal et~al.(2011)Agarwal, Reddy, R., and Rosé}]{SciSumm}
Nitin Agarwal, Ravi~Shankar Reddy, Kiran G.~V. R., and Carolyn~Penstein Rosé. 2011.
\newblock \href {http://dblp.uni-trier.de/db/conf/acl/acl2011d.html#AgarwalRRR11} {Scisumm: A multi-document summarization system for scientific articles.}
\newblock In \emph{ACL (System Demonstrations)}, pages 115--120. The Association for Computer Linguistics.

\bibitem[{Chang et~al.(2024)Chang, Lo, Goyal, and Iyyer}]{chang2024booookscore}
Yapei Chang, Kyle Lo, Tanya Goyal, and Mohit Iyyer. 2024.
\newblock \href {https://openreview.net/forum?id=7Ttk3RzDeu} {Booookscore: A systematic exploration of book-length summarization in the era of {LLM}s}.
\newblock In \emph{The Twelfth International Conference on Learning Representations}.

\bibitem[{Dagdelen et~al.(2024)Dagdelen, Dunn, Lee, Walker, Rosen, Ceder, Persson, and Jain}]{Dagdelen2024}
John Dagdelen, Alexander Dunn, Sanghoon Lee, Nicholas Walker, Andrew~S. Rosen, Gerbrand Ceder, Kristin~A. Persson, and Anubhav Jain. 2024.
\newblock \href {https://doi.org/10.1038/s41467-024-45563-x} {Structured information extraction from scientific text with large language models}.
\newblock \emph{Nature Communications}, 15(1):1418.

\bibitem[{Gunel et~al.(2024)Gunel, Wendt, Xie, Zhou, Vo, Fisher, and Tata}]{gunel2024strumllm}
Beliz Gunel, James~B. Wendt, Jing Xie, Yichao Zhou, Nguyen Vo, Zachary Fisher, and Sandeep Tata. 2024.
\newblock \href {http://arxiv.org/abs/2403.19710} {Strum-llm: Attributed and structured contrastive summarization}.

\bibitem[{Hwang et~al.(2024)Hwang, Zhou, Gunel, Wendt, and Tata}]{hwang2024sumie}
Eunjeong Hwang, Yichao Zhou, Beliz Gunel, James~Bradley Wendt, and Sandeep Tata. 2024.
\newblock \href {http://arxiv.org/abs/2406.05079} {Sumie: A synthetic benchmark for incremental entity summarization}.

\bibitem[{Jin et~al.(2024)Jin, Zhang, Meng, Wang, and Tan}]{jin2024comprehensive}
Hanlei Jin, Yang Zhang, Dan Meng, Jun Wang, and Jinghua Tan. 2024.
\newblock \href {http://arxiv.org/abs/2403.02901} {A comprehensive survey on process-oriented automatic text summarization with exploration of llm-based methods}.

\bibitem[{Kryscinski et~al.(2022)Kryscinski, Rajani, Agarwal, Xiong, and Radev}]{kryscinski2021booksum}
Wojciech Kryscinski, Nazneen Rajani, Divyansh Agarwal, Caiming Xiong, and Dragomir Radev. 2022.
\newblock \href {https://aclanthology.org/2022.findings-emnlp.488} {{BOOKSUM}: A collection of datasets for long-form narrative summarization}.
\newblock In \emph{Findings of the Association for Computational Linguistics: EMNLP 2022}, pages 6536--6558, Abu Dhabi, United Arab Emirates. Association for Computational Linguistics.

\bibitem[{Li et~al.(2024)Li, Zhang, Do, Yue, and Chen}]{li2024longcontext}
Tianle Li, Ge~Zhang, Quy~Duc Do, Xiang Yue, and Wenhu Chen. 2024.
\newblock \href {http://arxiv.org/abs/2404.02060} {Long-context llms struggle with long in-context learning}.

\bibitem[{Madaan et~al.(2022)Madaan, Tandon, Clark, and Yang}]{madaan2022memory}
Aman Madaan, Niket Tandon, Peter Clark, and Yiming Yang. 2022.
\newblock \href {https://aclanthology.org/2022.emnlp-main.183} {Memory-assisted prompt editing to improve {GPT}-3 after deployment}.
\newblock In \emph{Proceedings of the 2022 Conference on Empirical Methods in Natural Language Processing}, pages 2833--2861, Abu Dhabi, United Arab Emirates. Association for Computational Linguistics.

\bibitem[{Ouyang et~al.(2022)Ouyang, Wu, Jiang, Almeida, Wainwright, Mishkin, Zhang, Agarwal, Slama, Ray, Schulman, Hilton, Kelton, Miller, Simens, Askell, Welinder, Christiano, Leike, and Lowe}]{openai}
Long Ouyang, Jeffrey Wu, Xu~Jiang, Diogo Almeida, Carroll Wainwright, Pamela Mishkin, Chong Zhang, Sandhini Agarwal, Katarina Slama, Alex Ray, John Schulman, Jacob Hilton, Fraser Kelton, Luke Miller, Maddie Simens, Amanda Askell, Peter Welinder, Paul~F Christiano, Jan Leike, and Ryan Lowe. 2022.
\newblock \href {https://proceedings.neurips.cc/paper_files/paper/2022/file/b1efde53be364a73914f58805a001731-Paper-Conference.pdf} {Training language models to follow instructions with human feedback}.
\newblock In \emph{Advances in Neural Information Processing Systems}, volume~35, pages 27730--27744. Curran Associates, Inc.

\bibitem[{Xia et~al.(2024)Xia, Xing, Du, Yang, Feng, Xu, Yin, and Xiong}]{xia2024fofo}
Congying Xia, Chen Xing, Jiangshu Du, Xinyi Yang, Yihao Feng, Ran Xu, Wenpeng Yin, and Caiming Xiong. 2024.
\newblock \href {https://arxiv.org/abs/2402.18667} {Fofo: A benchmark to evaluate llms' format-following capability}.
\newblock \emph{ArXiv preprint}, abs/2402.18667.

\bibitem[{Xiong et~al.(2023)Xiong, Liu, Molybog, Zhang, Bhargava, Hou, Martin, Rungta, Sankararaman, Oguz, Khabsa, Fang, Mehdad, Narang, Malik, Fan, Bhosale, Edunov, Lewis, Wang, and Ma}]{long-context}
Wenhan Xiong, Jingyu Liu, Igor Molybog, Hejia Zhang, Prajjwal Bhargava, Rui Hou, Louis Martin, Rashi Rungta, Karthik~Abinav Sankararaman, Barlas Oguz, Madian Khabsa, Han Fang, Yashar Mehdad, Sharan Narang, Kshitiz Malik, Angela Fan, Shruti Bhosale, Sergey Edunov, Mike Lewis, Sinong Wang, and Hao Ma. 2023.
\newblock \href {http://arxiv.org/abs/2309.16039} {Effective long-context scaling of foundation models}.

\bibitem[{Zhang et~al.(2023)Zhang, Zhao, Kang, and Liu}]{zhang2023memory}
Kai Zhang, Fubang Zhao, Yangyang Kang, and Xiaozhong Liu. 2023.
\newblock \href {https://arxiv.org/abs/2309.11696} {Memory-augmented llm personalization with short-and long-term memory coordination}.
\newblock \emph{ArXiv preprint}, abs/2309.11696.

\end{thebibliography}
\bibliographystyle{acl_natbib}

\appendix

\newpage 

\section{Number of tokens under limited token size scenario}
\label{app:sumie-compress-tokens}
\begin{figure}[t]
    \small
    \includegraphics[width=0.8\linewidth]{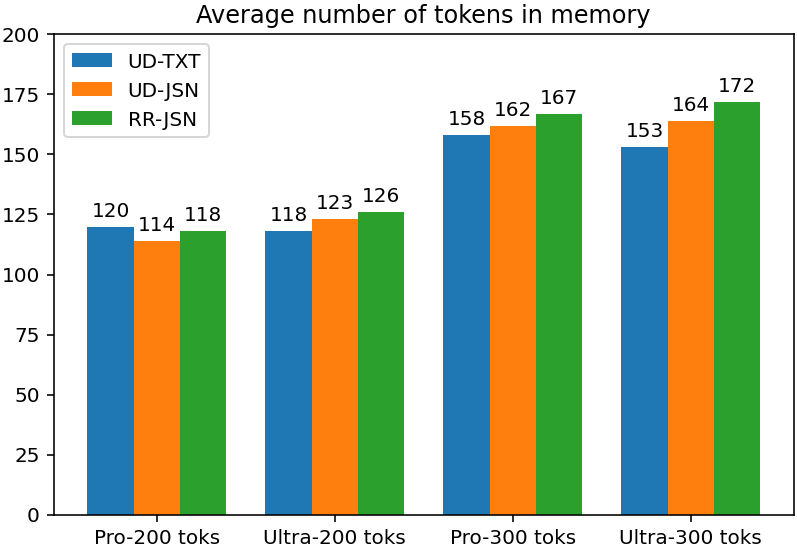}
    \caption{Average number of tokens used as an existing information across all turns with limited memory token size on SUMIE.}
    \label{fig:sumie-compress-tokens}
\end{figure}

Figure \ref{fig:sumie-compress-tokens} shows the average number of tokens used for existing information in prompts across all turns. We observe that the models effectively compress tokens across all methods. Specifically, when the token limit is set to 200 tokens, the models compress the information to fewer than 130 tokens, and when the limit is set to 300 tokens, they compress it to fewer than 180 tokens.

\section{Predefined schema for entity summarization and book summarization tasks}
\label{app:schema}
For entity summarization, we define the schema as follows:

\begin{framed}
\tiny
\tt
Summary: $\{$attributes: $\{$name: str, values: list[str]$\}\}$
\end{framed}

To construct structured representations of book contexts, we define the schema as follows:

\begin{framed}
\tiny
\tt
Summary:  $\{$\\
    characters:  $\{$
        name: str,
        explanations: list[str]
    $\}$,\\
    events:  $\{$
        name: str,
        explanations: list[str]
    $\}$,\\
    background: $\{$
        name: str,
        explanations: list[str]
    $\}$,\\
    motivations: $\{$
        name: str,
        explanations: list[str]
    $\}$,\\
    objectives: $\{$
        name: str,
        explanations: list[str]
    $\}$,\\
    other: $\{$
        name: str,
        explanations: list[str]
    $\}$
$\}$
\end{framed}

\section{Details about Dataset}
\label{app:dataset}
SUMIE contains 200 entities and each entity is associated with 7 paragraphs. Each paragraph has aggregated a summary, which includes aggregated attribute-value pairs from the 1st paragraph to $N$-th paragraph (e.g. 7th paragraph contains aggregated attribute-value pairs from all 7 paragraphs). Each summary contains attribute-value pairs that are relevant to the specific entity. BookScore dataset contains mostly fiction or comtemporary books. Refer to \citet{chang2024booookscore} for the full book list.

\section{Details about Evaluation}
\label{app:eval-details}
On SUMIE, the evaluation of the final summary comprises attribute-value pairs related to a specific entity, measuing precision, recall, and f1 score. We use \pro as an LLM evaluator. The temperature is set to 0.8.

For BookScore dataset, we use their LLM-based evaluation metric that assess summary coherence based on 8 predefined error dimensions (entity omission, event omission, causal omission, discontinuity, salience, language, inconsistency, duplication).
Each error measures the following aspects:

\begin{figure}
\small
\begin{framed}
\tt
- Entity omission: an entity, real or abstract (person, object, place, concept, etc.) is mentioned, but key details are missing or unclear\\
- Event omission: an event is mentioned, but key details are missing or unclear\\
- Causal omission: the reason or motivation for something is missing or unclear\\
- Salience: inclusion of trivial details that do not contribute to the main storyline\\
- Discontinuity: an interruption in the flow of the narrative, including but not restricted to: sudden jumps between perspectives, time periods, or settings; poor transition between sentences or paragraphs; sentences or paragraphs that seem out of place; illogical sentence order or summary structure\\
- Duplication: redundant repetition of similar information\\
- Inconsistency: two parts of the summary contain contradicting information\\
- Language: grammar issues; confusing wording or phrasing; etc.
\end{framed}
\end{figure}

To evaluate the summary, each sentence is broken down into individual sentences and given to an LLM evaluator along with the original summary. The evaluation prompt includes multiple examples for each error dimension, and the LLM is asked to determine whether any errors are present in the current sentence and summary. If any errors are found, the sentence is marked as a ``confusing sentence''. The final score is calculated by dividing the total number of confusing sentences identified by the LLM by the total number of sentences in the summary.

Since their evaluation requires an advanced ability of understanding book context to identify 8 predefined error types, we use \ultra as an LLM evaluator and the temperature was set to 0.8. To calculate the token size to create a list of smaller book chunks, we use \texttt{tiktoken} library\footnote{https://github.com/openai/tiktoken}.
We used TPU v5e for evaluating the entity summarization and book summarization tasks, with each evaluation taking up to 24 hours. In particular, Gemini-Pro used 8 pods and Gemini-Ultra used 64 pods. All experimental results are based on a single run using the prompts provided in each section.

\section{Markdown vs. JSON}
\label{app:markdown}
In our evaluation, presented in Table \ref{tab:eval-results-markdown}, we compared the effectiveness of JSON and Markdown in data structuring for summarization tasks. While both formats demonstrated comparable abilities in organizing straightforward data with the Gemini Pro model, JSON distinguished itself in handling more complex scenarios. Its capability to support nested dictionary structures enhances expressibility and summarization precision, particularly in lengthy or hierarchically complex documents. This makes JSON especially valuable for summarizing detailed datasets like family trees, where its hierarchical structuring capabilities far outperform the linear layout of Markdown.

\begin{table}[t]
\scriptsize
\centering
\begin{tabular}{l|l|ccc}
\toprule

Method & Turn & P & R & F1 \\
\midrule
GM\textsubscript{markdown} & last & 83.1 & 74.8 & 78.3 \\
 & Avg. & 80.8 & 80.8 & 80.4 \\ 
\midrule

GM\textsubscript{json} & last & 84.6 & 74.3 & 78.6 \\
 & Avg. & \textbf{84.7} & 78.7 & 80.9 \\ 

\midrule
CoK\textsubscript{json} & last & 84.6 & 83.6 & 83.9 \\ 
 & Avg. & 83.7 & \textbf{83.9} & \textbf{83.5} \\

\bottomrule
\end{tabular}

\caption{Performance on SUMIE with \pro on Generate-Merge (GM) and Chain-of-Key (CoK) update with markdown and json formats.}
\label{tab:eval-results-markdown}
\end{table}

\section{Compressing the information}
\label{app:compress-prompts}
To manage the token size for existing information, we compress the content using 3 criteria with an LLM when the token size exceeds $K$ tokens: 1) Redundancy, 2) Frequency, and 3) Relevance. For redundancy, we remove repetitive information to maintain conciseness. For frequency, the model prioritizes the most frequently mentioned values, as they are likely the most important. For relevance, the model emphasizes information most pertinent to the subject. Figure \ref{fig:prompt-compress} presents the prompt used for compressing the information.

\section{Prompt for Chain-of-Key update}
\label{app:cok-prompts}
Figure \ref{fig:cok-prompt} is the prompt used for Chain-of-Key update process. In the case of the book summarization task, we simply replace the instructions and a example given in the prompt:

\begin{framed}
\tt
\small
1. Values have a short and concise information: the values of the [PARTIAL  SUMMARY] should have a short, concise, and summarized information.\\
2. No redundant keys: If information from [NEW  SUMMARY] can be incorporated by updating an existing key in [PARTIAL  SUMMARY], then do not introduce a new redundant key. For example, if there's already a field for 'activities' do not introduce a new key for 'other activities' or 'water activities', 'hiking'. Update the existing key for 'activities'.\\
3. No redundant values under the same key: If one value encompasses most of the details in another value, merge them together. For instance, "beautiful views of the Eiffel tower" and "view of the Eiffel tower" should be merged into a single value like "beautiful views of the Eiffel tower\\
4. Do not include trivial information or redundant information as a value for its corresponding key.\\
5. Content Focus: Values should highlight the most important information relevant to the main story.\\
6. Exclude Ancillary Content: Disregard sections that are not directly part of the main narrative, such as: Title, Acknowledgments, Dedication, Chapter titles, Glossary entries, Timelines, Forewords, Prologues, Epilogues, Appendices, Author notes.
\end{framed}

\begin{figure*}[ht]
\begin{framed}
\tt
\tiny
I will provide a JSON format summary in a section called [NEW SUMMARY], and a class definition [CLASS], which define some fields that need to be generated, and an instantiation of that class under [PARTIAL SUMMARY] that is a response to the question in [QUESTION]. Your task is to propose updates to [PARTIAL SUMMARY] gathered from the information in [NEW SUMMARY].\\

There are two types of revisions that you can suggest: ADD and UPDATE.\\

For UPDATE, follow these instructions:\\
1. Your proposed updates must be for valid JSONPaths that already exist in [PARTIAL SUMMARY]. If the JSONPath does not exist, you should not propose an update for that JSONPath.\\
2. Updates can be made by modifying an existing value using content from [NEW SUMMARY].\\
3. Updates should never reduce the amount of information in [PARTIAL SUMMARY].\\
4. Never remove existing information from the [PARTIAL SUMMARY].\\
4. Proposed update must be a `dict[str, ProposedUpdate]` where the key is a valid JSONPath in [CLASS] and `ProposedUpdate` is defined as follows:\\
```\\
class ProposedUpdate(TypedDict):\\
  update: Any  \# The type must be the same type as at the JSONPath in [CLASS].\\
```\\

For ADD, follow these instructions:\\
1. Proposed additions must be for valid JSONPaths that adhere to the definition in [CLASS]. They are allowed to increase the size of lists in the definition, but they must not define new fields which are not defined in the class definition.\\
2. It is OK to add partial objects. Leave fields unset if [NEW SUMMARY] does not contain a value for one of the fields in [PARTIAL SUMMARY].\\
3. Proposed additions must be a `dict[str, ProposedAdd]` where the key is a valid JSONPath in [CLASS] and `ProposedAdd` is defined as follows:\\
```\\
class ProposedAdd(TypedDict):\\
  add: Any  \# The type must be the same type as at the JSONPath in [CLASS].\\
```\\

For both operations, follow these instructions:\\
1. Values have sufficient context: the values of the [PARTIAL SUMMARY] should have enough context so a reader can understand what it means.\\
2. No redundant keys: If information from [NEW SUMMARY] can be incorporated by updating an existing key in [PARTIAL SUMMARY], then do not introduce a new redundant key.\\
3. No redundant values under the same key: If one value encompasses most of the details in another value, merge them together.\\

[QUESTION]\\
Merge the new summary and existing summary of HOTEL0.\\

[NEW SUMMARY]\\
{\\
  "attributes": {\\
    "Room Amenities": ["pub opens till midnight", "two large pools"],\\
    "Noise Level": ["Notable street noise at night"],\\
  }\\
}\\

[CLASS]\\
class Summary(TypedDict):\\
  attributes: dict[str, list[str]]  \# Keyed by attribute, with a list of sufficient details about the attribute.\\

[PARTIAL SUMMARY]\\
{\\
  "attributes": {\\
    "Amenities": ["two pools"],\\
    "Food \& Beverage": ["limited breakfast options"],\\
  }\\
}\\

[THOUGHTS FOR UPDATE]\\
1. I need to figure out which fields and values to update.\\
2. [PARTIAL SUMMARY] contains information about the following: ["Amenities", "Food \& Beverage"]\\
3. [NEW SUMMARY] contains new content relevant to the following existing content: ["Amenities"]\\
4. The content should be updated at the following JSONPaths: ["$\$$.'attributes'.'Amenities'"]\\

[UPDATED OBJECTS]\\
$\{$\\
  "$\$$.'attributes'.'Amenities'": $\{$"update": [ "pub opens till midnight" ]$\}$\\
$\}$\\

[THOUGHTS FOR ADD]\\
1. I need to figure out which fields and values to add.\\
2. [NEW SUMMARY] mentions information about the following: ["Amenities", "Noise Level"]\\
3. [PARTIAL SUMMARY] does not yet have information about: [ "Noise Level" ]\\
3. The content should be added at the following JSONPaths: [ "$\$$.'attributes'.'Noise Level'"]\\

[ADDED OBJECTS]\\
$\{$\\
  "$\$$.'attributes'.'Noise Level'": $\{$"add": [ "Notable street noise at night" ]$\}$,\\
$\}$\\
\end{framed}
\caption{Prompt used for chain-of-key update}
\label{fig:cok-prompt}
\end{figure*}

\section{Prompts used for SUMIE baselines}
\label{app:sumie-basline-prompts}
We used Figure \ref{fig:prompt-generate} for Generate-Once, Figure \ref{fig:prompt-generate} and \ref{fig:prompt-update} for Generate-Update, and Figure \ref{fig:prompt-generate} and \ref{fig:prompt-remove-dup} are used for Generate-Merge, which includes removing duplicates. For text baselines, we simply replace the JSON examples in the prompts to text summary examples.

\begin{figure*}[ht]
\small
\begin{framed}
\tt
Task Overview:\\
Your task involves synthesizing information from detailed descriptive paragraphs about a specific entity into a summary table.\\
This Json will highlight key attributes of the entity along with their detailed descriptions derived from the given texts.\\

Instructions:\\
* Extract Descriptive Values: Focus on extracting specific, detailed information rather than general or vague adjectives like "good" or "bad." Ensure that descriptions are precise and informative.\\
* Present a Balanced View: The table should reflect a balanced perspective, including positive, negative, and neutral attributes. For attributes with mixed reviews, indicate the sources supporting each viewpoint.\\
* Attribute Selection:\\
 - Commonly Interested Attributes: Include attributes that are generally of interest for the type of entity being described.\\
 - Unique Attributes: Also identify and include unique attributes that are specifically mentioned in the provided descriptions.\\
* Do not include irrelevant sentences about the given entity in the summary. Irrelevant sentences include entity names (HOTEL1, HUMAN) that are different from the given entity (HOTEL0).\\

Structure of the Summary Table:\\
* The Json should contain a dictionary format data, where keys are attributes and values are detailed descriptions of their corresponding attributes.\\
* List attributes with their corresponding values, indicating the source paragraph and relevant excerpts for substantiation.\\
* If an attribute has multiple values, include all values as a list of the attribute.\\
* Each value should contain sufficient evidence extracted from the paragraph related to the entity.\\

Example:\\
Entity: HOTEL0\\

Paragraphs:\\
P1. Great room and service, but breakfast was lacking. We loved the spacious room and friendly staff, but the breakfast options were limited. There are two pools.\\
P2. Poor customer service overshadowed the beautiful location. The beachfront view was amazing, but dealing with unhelpful staff was frustrating. Room is comfortable.\\
P3. Exceptional dining and comfortable beds, but noisy at night. The restaurant was five-star, and the beds were very cozy, but there was a lot of street noise.\\
P4. HOTEL1 offers great room service and breakfast was amazing. (Irrelevant sentence for the given entity "HOTEL0")\\
P5. HUMAN's creativity looks like a great room service offered by the hotel. (Irrelevant sentence for the given entity "HOTEL0")\\

Summary JSON:\\
$\{$
  "Room Quality": ["Spacious and comfortable rooms"],\\
  "Amenities": ["There are two pools"],\\
  "Service": ["Friendly staff", "overshadowed by unhelpful staff"],\\
  "Location": ["Beautiful beachfront view"],\\
  "Food \& Beverage": ["Exceptional dining experience", "limited breakfast options"],\\
  "Noise Level": ["Notable street noise at night"]\\
$\}$

Your Task:\\
Generate a similar table based on the following descriptions of the specified entity.
Entity: $\{entity\_name\}$\\

Paragraphs:\\
$\{paragraph\}$

Proceed to generate the summary Json.

\end{framed}
\caption{Prompt used for generating initial summary}
\label{fig:prompt-generate}
\end{figure*}

\begin{figure*}[ht]
\small 
\begin{framed}
\tt
Task Overview:
You are tasked with refining and expanding an existing summary table based on new descriptive paragraphs about an entity.\\
This involves updating the table to include new information, modify existing details without removing any, and ensuring all entries are supported by evidence from the text.\\

Instructions:\\
* Update Descriptive Values: Carefully read the new paragraph(s) and identify any information that should be added to the current table entries or modify them. Focus on specific, descriptive details, avoiding vague adjectives. **Do not remove any existing attributes or values**, but rather add to or revise them as necessary.\\
* Maintain a Balanced View: Ensure the updated table continues to present a balanced perspective, incorporating positive, negative, and neutral values. For any attribute with mixed evidence, update the count of sources supporting each view. All original attributes and values must be preserved in the table, with modifications only to reflect new insights or corrections based on the latest information.\\
* Attribute Revision and Addition:
 - Commonly Interested Attributes: Update or add attributes that are of general interest for the type of entity being described, based on the new information.
 - Unique Attributes: Identify and incorporate any unique attributes mentioned in the new paragraphs that were not previously included in the table.\\

Structure of the Updated Summary Table:\\
* Retain the two-column format: Attribute and Value.\\
* For each attribute, list the updated or new evidence indicating the source paragraph and relevant excerpts. Original attributes and values should remain listed, with additional information appended as necessary.\\
* If an attribute has multiple values, include all values as a list of the attribute.\\
* Each value should contain sufficient evidence extracted from the paragraph related to the entity.\\

Example\\
Entity: Hotel0\\
New Paragraph:\\
P4. The hotel has recently renovated its lobby, which now features a modern design.  Guests have also noted improvements in breakfast variety and quality.\\
P5. The hotel boasts impeccably designed rooms, featuring luxurious furnishings.\\

Given Existing Summary Table:\\
$\{$\\
  "Room Quality": ["Spacious and comfortable rooms"],\\
  "Amenities": ["two pools"],\\
  "Service": ["Friendly staff", "overshadowed by unhelpful staff"],\\
$\}$\\

Updated Summary Json:\\
$\{$\\
  "Room Quality": ["Spacious and comfortable rooms", "Impeccably designed", "luxurious furnishings"],\\
  "Amenities": ["Two pools"],\\
  "Service": ["Friendly staff", "overshadowed by unhelpful staff"],\\
  "Food \& Beverage": ["Exceptional dining experience", "limited breakfast options", "improved breakfast variety and quality"],\\
  "Lobby Design": ["Modern design"],\\
$\}$\\

Your Task:\\
Update the summary Json with the given new descriptions of the specified entity.\\
Entity: $\{entity\_name\}$\\
New Paragraph:\\
$\{paragraph\}$\\

Given Existing Summary Json:\\
$\{existing\_summary\}$\\

Proceed to update the summary Json.

\end{framed}
\caption{Prompt used for updating a summary with new information and existing summary information.}
\label{fig:prompt-update}
\end{figure*}

\begin{figure*}[ht]
\small
\begin{framed}
\tt
Task Overview:\\
Your task involves removing duplicate information from a detailed summary json about a specific entity. This summary will highlight key attributes of the entity along with their detailed descriptions derived from the given texts.\\

Instructions:\\
1. Eliminate repetitive information to ensure the summary is concise.\\
2. In the given summary json, the keys are attributes of the entity and each attribute has its corresponding values.\\
3. If one attribute encompasses most of the details in another attribute, merge them together.\\
4. If one value encompasses most of the details in another value, merge them together.\\

Here is an example of merging attributes:

Given Existing Summary:\\
$\{$\\
    "Views": ["beautiful views of the Eiffel tower"],\\
    "views from hotel": ["visible Eiffel tower"],\\
$\}$\\

New Summary after removing duplicates and merging:\\
$\{$\\
    "View": ["beautiful views of the Eiffel tower"]\\
$\}$\\

===

Here is an example of merging values:

Given Existing Summary:\\
$\{$\\
    "Views": ["beautiful views of the Eiffel tower", "view of the Eiffel tower"],\\
    "views from hotel": ["visible Eiffel tower"],\\
$\}$\\

New Summary after removing duplicates and merging:\\
$\{$\\
    "View": ["beautiful views of the Eiffel tower"]\\
$\}$\\

===

\end{framed}
\caption{Prompt used for removing duplicates.}
\label{fig:prompt-remove-dup}
\end{figure*}

\section{Prompts used for BookScore baselines}
\label{app:book-basline-prompts}
For the BookScore dataset, we used the prompt in Figure \ref{fig:prompt-generate-book} along with special instructions from Figure \ref{fig:prompt-json-instruction} for JSON format generation and Figure \ref{fig:prompt-text-instruction} for plain text summary generation during the initial Generate-Update phase and the Generate-Merge phase. In subsequent Generate-Update iterations, we used the prompt in Figure \ref{fig:prompt-update-book}. To remove duplicates during the Generate-Merge step, we used the prompt in Figure \ref{fig:prompt-remove-dup}.

\begin{figure*}[ht]
\small
\begin{framed}
\tt
Task Overview:\\
We are analyzing segments of a story sequentially to progressively build a comprehensive summary of the entire plot. Your task is to generate a new summary by integrating vital information from the current story segment with the existing summary stored in memory. The summary can be provided in either text format or JSON format.\\

Instructions:\\
1. Integrate Key Information: Incorporate new information related to key events, backgrounds, settings, characters, their objectives, and motivations from the current segment into the existing summary.\\
2. Introduction of New Elements: Briefly introduce any new characters, places, or major elements mentioned for the first time in the current segment if they are not already included in the memory.\\
3. Handling Non-Linear Narratives: Account for non-linear narratives, including flashbacks, and switches between alternate worlds or viewpoints, ensuring the summary maintains a consistent and chronological narrative.\\
4. Cohesive Summary: Create a summary that reads as though it was written in one go, despite the step-by-step process of updating it with each new segment.\\
5. Exclude Ancillary Content: Disregard sections that are not directly part of the main narrative, such as: Title, Acknowledgments, Dedication, Chapter titles, Glossary entries, Timelines, Forewords, Prologues, Epilogues, Appendices, Author notes.\\

$\{special\_instruction\}$

Your Task:\\
Generate a summary based on the following segment from a story and the memory of the story up until this point. Ensure the output follows the specified format.

A segment from a story:

---

$\{book\_chunk\}$

---

Generated summary in $\{output\_format\}$:

\end{framed}
\caption{Prompt used for generating book summaries.}
\label{fig:prompt-generate-book}
\end{figure*}

\begin{figure}[ht]
\small
\begin{framed}
\tt
Structure of the JSON Summary:\\
- Fields to Generate: Characters, Events, Backgrounds, Motivations, Objectives, Other.\\
- Field Format: Each field should be a dictionary where keys are the names of elements and values are their short descriptions.\\
- Each key should include a short and concise information as values that explain the key.\\
- Content Focus: Values should highlight the most important information relevant to the main story.\\
- Do not include trivial information or redundant information as a value for its corresponding key.\\

Here is an example of the JSON Summary:\\
$\{$\\
  "characters": $\{$\\
    "a character's name": [a list of short and summarized descriptions]\\
  $\}$,\\
  "events": $\{$\\
    "an event's name": [a list of short and summarized descriptions]\\
  $\}$,\\
  "objectives": $\{$\\
    "an objective's name": [a list of short and summarized descriptions]\\
  $\}$,\\
  "motivations": $\{$\\
    "a motivation's name": [a list of short and summarized descriptions]\\
  $\}$,\\
  "background": $\{$\\
    "a background's name": [a list of short and summarized descriptions]\\
  $\}$,\\
  "other": $\{$\\
    "other information's name": [a list of short and summarized descriptions]\\
  $\}$\\
$\}$\\

\end{framed}
\caption{Instructions used for generating JSON format summary.}
\label{fig:prompt-json-instruction}
\end{figure}

\begin{figure}[ht]
\small
\begin{framed}
\tt
Structure of the Text Summary:\\
- Key Elements to Include: Incorporate key events, characters, backgrounds, motivations, objectives, and other relevant details.\\
- Narrative Flow: Ensure the summary flows seamlessly as a cohesive and comprehensive narrative.\\

Here is an example of the Text Summary format:\\
A summary that reads as though it was written in one go. It can consist of multiple paragraphs.\\

\end{framed}
\caption{Instructions used for generating text format summary.}
\label{fig:prompt-text-instruction}
\end{figure}

\begin{figure}[ht]
\small
\begin{framed}
\tt
Your Task: \\
Generate a summary based on the following segment from a story and the memory of the story up until this point. Ensure the output follows the specified format.\\

A segment from a story:\\

---\\

$\{book\_chunk\}$

---\\

A memory of the story up until this point:\\

---

$\{memory\}$

---\\

Output Type: $\{output\_format\}$\\

Updated summary in $\{output\_format\}$:\\

\end{framed}
\caption{Prompt used for updating a summary with a new information and existing summary.}
\label{fig:prompt-update-book}
\end{figure}

\begin{figure}[ht] 
\begin{framed}
\tt
\small
Task Overview:
Your task involves compressing information from a detailed summary JSON about a book. This summary will highlight key details of the book that are important when summarizing the whole story of the book. \\

Instructions: \\
- Compress the summary to the specified number of tokens below. \\
- The condensed summary should retain key details about characters, events, backgrounds, motivations, objectives, and other important information. \\
- If the key has multiple values, merge them into a short summarized description. \\

Criteria: \\
- Redundancy: Eliminate repetitive information to ensure the summary is concise. \\
- Frequency: Emphasize the most frequently mentioned attributes or values, as they are likely the most important. \\
- Relevance: Focus on the information that is most pertinent to the main story of the book or the overall context of the summary. \\
- Remove trivial information that does not frequently appear in the other contexts or not relevant to the main story of the book based on the overall context of the summary. 
\end{framed}
\caption{Prompt used to compress the information to fit the existing summary into the limited token size.}
\label{fig:prompt-compress}
\end{figure}
\label{sec:appendix}


\end{document}